\documentclass[10pt,twocolumn,letterpaper]{article}

\usepackage[pagenumbers]{iccv} % To force page numbers, e.g. for an arXiv version

\usepackage{cuted}  
\definecolor{iccvblue}{rgb}{0.21,0.49,0.74}

\def\paperID{10995} % *** Enter the Paper ID here
\def\confName{ICCV}
\def\confYear{2025}
\usepackage[pagebackref,breaklinks,colorlinks,allcolors=iccvblue]{hyperref}

\title{PSF-4D: A Progressive Sampling Framework for View Consistent 4D Editing}

\author{Hasan Iqbal$^{*,1}$, Nazmul Karim$^{*,2}$, Umar Khalid$^{2}$, Azib Farooq$^{3}$, Zichun Zhong$^{1}$ \\  Chen Chen$^2$, Jing Hua$^{1}$ \\
  \small{$^{1}$Wayne State University}
  \small{$^{2}$University of Central Florida} 
  \small{$^{3}$Miami University}\\
  {\tt\small nazmul.karim170@gmail.com, \{hasan.iqbal.cs, zichunzhong, jinghua\}@wayne.edu}\\ 
  {\tt\small umar.khalid@ucf.edu, azib.farooq@miamoh.edu, chen.chen@crcv.ucf.edu} \vspace{0.5mm} \\
}

\begin{document}
\maketitle

% \institute{University of Central Florida, Orlando, FL, USA \and
% Department of Computer Science, Wayne State University, Detroit, MI, USA}
\begingroup
\renewcommand\thefootnote{}
\footnotetext{$^*$Equal contribution}
\endgroup

\begin{strip}
\centering
\includegraphics[width=0.95\linewidth]{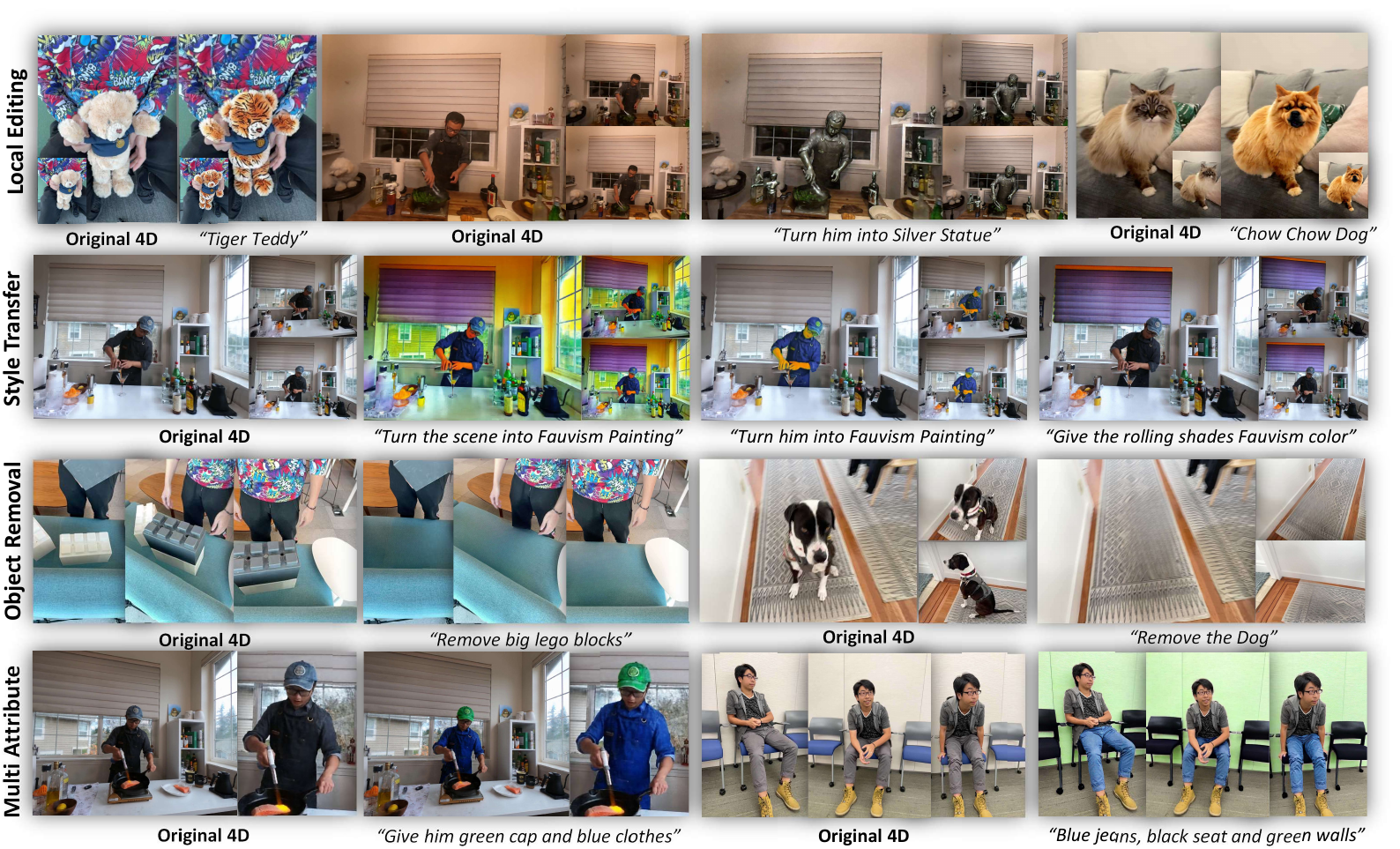}
    \captionof{figure}{\footnotesize Examples of 4D editing tasks with our approach, covering \textbf{Local Editing}, \textbf{Style Transfer}, \textbf{Object Removal}, and \textbf{Multi-Attribute Editing}. \textbf{Local Editing}: Transforms objects like a teddy bear into a \textit{Tiger Teddy} and a cat into a \textit{Chow Chow Dog}. \textbf{Style Transfer}: Applies artistic styles, such as \textit{Fauvism Painting}, across frames, maintaining visual coherence. \textbf{Object Removal}: Eliminates objects (e.g., \textit{big lego blocks, dog}) while preserving background consistency. \textbf{Multi-Attribute Editing}: Combines edits, such as changing hair to blue and clothing to silver, or adding attributes like "Blue jeans, black seat, and green walls. These examples demonstrate our model’s ability to perform complex 4D edits with spatial and temporal consistency across various scenarios.} 
    \label{fig:teaser}
\label{fig:teaser}
\end{strip}
\begin{abstract}
Instruction-guided generative models, especially those using text-to-image (T2I) and text-to-video (T2V) diffusion frameworks, have advanced the field of content editing in recent years. To extend these capabilities to 4D scene, we introduce a progressive sampling framework for 4D editing (PSF-4D) that ensures temporal and multi-view consistency by intuitively controlling the noise initialization during forward diffusion.  For temporal coherence, we design a correlated Gaussian noise structure that links frames over time, allowing each frame to depend meaningfully on prior frames. Additionally, to ensure spatial consistency across views, we implement a cross-view noise model, which uses shared and independent noise components to balance commonalities and distinct details among different views. To further enhance spatial coherence, PSF-4D incorporates view-consistent iterative refinement, embedding view-aware information into the denoising process to ensure aligned edits across frames and views. Our approach enables high-quality 4D editing without relying on external models, addressing key challenges in previous methods. Through extensive evaluation on multiple benchmarks and multiple editing aspects (e.g., style transfer, multi-attribute editing, object removal, local editing, etc.), we show the effectiveness of our proposed method. Experimental results demonstrate that our proposed method outperforms state-of-the-art 4D editing methods in diverse benchmarks.
\end{abstract}  
% We further enforce view awareness in the video diffusion model using a cross-view attention mechanism where we infuse information of one view while denoising another view.    
\vspace{-1.5mm}
\section{Introduction}\label{sec:intro}
% Instruction-guided content generation~\cite{pnp,p2p,ruiz2023dreambooth,singer2023makeavideo,zhou2022magicvideo} has gained momentum in the last couple of years due to the high utility of content editing in diverse real-world applications. The most common instruction is "text prompt" while diffusion models became the go-to technique for accommodating generation tasks. Among these generative models, text-to-image (T2I)~\cite{pix2pix-zero, brooks2023instructpix2pix,couairon2023diffedit} and text-to-video (T2V)~\cite{liu2023video, ceylan2023pix2video, qi2023fatezero} editing models have been heavily explored with continuous efforts still taking place.   Recently, several approaches using T2I diffusion models have been applied to 3D scene generation and editing~\cite{kamata2023instruct,haque2023instruct, brooks2023instructpix2pix}, combining a neural 3D representation such as NeRF with an image diffusion model. For example, InstructNeRF2NeRF~\cite{haque2023instruct} facilitates text-driven 3D scene editing by iteratively aligning the rendered views from the 3D model with those produced by the diffusion model, enabling flexible edits via textual input. In our work, we explore the paradigm of 4D scene editing using a T2I diffusion model.

Instruction-guided content generation~\cite{pnp,p2p,ruiz2023dreambooth,singer2023makeavideo,zhou2022magicvideo} has seen rapid advancements, propelled by the the effectiveness of diffusion models across various domains. Among these, text-to-image (T2I)~\cite{pix2pix-zero, brooks2023instructpix2pix,couairon2023diffedit} and text-to-video (T2V)~\cite{liu2023video, ceylan2023pix2video, qi2023fatezero, karim2023save} generation have garnered significant attention, enabling high-fidelity synthesis and manipulation. Building on these successes, recent efforts have extended T2I diffusion models to 3D scene editing~\cite{kamata2023instruct,haque2023instruct, karim2024free, khalid2023latenteditor, brooks2023instructpix2pix}, integrating image diffusion with neural 3D representations such as NeRF to facilitate flexible, text-driven modifications. In this work, we take a step further by exploring 4D scene editing, leveraging a T2I diffusion model to enable temporally consistent and semantically meaningful scene transformations.  

The field of 4D scene reconstruction has witnessed significant advancements with the development of dynamic neural 3D representations~\cite{fridovich2023k, liu2022devrf}, including K-Planes~\cite{kplanes}, HexPlanes~\cite{hexplane}, and dynamic 3D Gaussian fields~\cite{luiten2024dynamic, wu20244d}. These methods have substantially improved our ability to capture and model temporally evolving scenes with high fidelity. Conceptually, a 4D scene can be viewed as a pseudo-3D representation~\cite{mou2024instruct}, where each viewpoint corresponds to a video rather than a static image. Consequently, adapting a text-to-image (T2I) model for 4D scene editing necessitates extending it into a text-to-video (T2V) framework.  

However, ensuring temporal and multiview consistency in edits remains a key challenge. Variations in modifications across different viewpoints and time frames can introduce significant inconsistencies, complicating interactive 4D scene editing. Recent approaches, such as Control4D~\cite{shao2023control4d} and Instruct-4D-to-4D~\cite{mou2024instruct}, have made strides toward addressing these challenges, but they rely heavily on auxiliary models beyond diffusion-based architectures. For example, Instruct-4D-to-4D employs a pre-trained optical flow model~\cite{teed2020raft} to enforce consistency, while Control4D integrates a GAN-based refinement module. These dependencies introduce inherent limitations: GAN training can be unstable, optical flow models may struggle in complex or unseen scenarios, and Instruct-4D-to-4D’s anchor-aware attention mechanism can lead to inconsistencies depending on anchor selection. In this work, we aim to overcome these limitations by leveraging the internal forward and reverse sampling processes of diffusion models, ensuring a more principled and end-to-end diffusion-based approach for 4D scene editing.  

Building on these insights, we present \textbf{PSF-4D}, a novel 4D editing framework that introduces \textit{progressive noise sampling} and \textit{iterative refinement} to enhance generation quality. Prior works~\cite{lee2021priorgrad, shi2023mvdream, ge2023preserve, luo2023videofusion} have demonstrated that careful noise control and multiview geometry information can significantly improve diffusion-based synthesis. Inspired by this, we propose a targeted manipulation of noise initialization during the forward diffusion phase, coupled with view-consistent noisy latent refinement in the reverse diffusion phase. To ensure temporal coherence, we leverage the autoregressive nature of temporal data by explicitly modeling relationships across the sequence.  

However, robust 4D editing demands not only temporal consistency but also view consistency across perspectives. To address this, we introduce a \textbf{cross-view noise model} within the Text-to-Video (T2V) framework, enhancing spatial alignment across views. CNM builds upon principles of 3D multiview geometry, enforcing spatial coherence by decomposing noise into two complementary components: a \textit{shared component} that captures cross-view similarity and an \textit{independent component} that preserves view-specific variations. While noise initialization plays a key role in maintaining coherence, it alone is insufficient to enforce consistency across edits. To this end, we develop a \textbf{view-consistent iterative refinement} mechanism that directly integrates view-aware editing signals into the denoising stages of the diffusion model. This strategy enforces consistent modifications across perspectives while retaining necessary view-dependent details, ensuring both temporal and spatial coherence in the final 4D output. Our key contributions are summarized below:

\begin{itemize}
    \item We introduce several straightforward yet impactful modifications to the core diffusion process of a text-to-video model, leveraging progressive noise sampling and iterative latent refinement techniques. 
    % These enhancements enable us to achieve robust 4D scene editing with both temporal and view consistency.
    
    \item By intuitively controlling noise in the diffusion process, we establish coherence across noisy video frames captured from different views, which leads to 4D generation with reduced inconsistencies. A refinement strategy focusing solely on improving view consistency is introduced to further refine the edited 4D model. 
    
    \item Through comprehensive evaluations across various benchmarks and diverse editing tasks, we demonstrate the effectiveness of PSF-4D as shown in Figure~\ref{fig:teaser}. 
\end{itemize}

% In addition, with simple modifications to relevance-map guided editing~\cite{} we can achieve local editing capabilities by accurately estimating the contribution of each pixel to the desired edit.  

%  Ultimately, this approach aims to improve the quality and fidelity of generated video or time-series data by leveraging time-based relational patterns. This design enhances the temporal coherence and consistency across generated frames, leading to smoother transitions and more realistic sequences in the final output. We propose a noise design aimed at improving the temporal consistency of generated data. Instead of using standard Gaussian noise during both training and inference stages, we introduce autoregressive-structured correlations for the noises across time. This targeted approach enables a more cohesive and coherent generation process. By introducing this structured noise into the diffusion process, we achieve enhanced temporal coherence across frames in the generated video. This autoregressive correlation structure allows each frame’s noise to depend on the noise from previous frames, promoting smoother and more temporally consistent transitions in the generated sequences.
\section{Related Work}
% \vspace{-1cm}
\paragraph{Diffusion-Based Video Editing.}
Diffusion-based generative models have excelled in text-guided image editing~\cite{ruiz2023dreambooth, couairon2023diffedit, pnp, p2p, sdedit, pix2pix-zero, brooks2023instructpix2pix}, but adapting them for video editing presents unique challenges, especially in preserving temporal coherence. A common approach is to transform Text-to-Image (T2I) models into Text-to-Video (T2V) models. For instance, Tune-A-Video~\cite{wu2023tune} adds temporal self-attention layers for one-shot fine-tuning, while Make-A-Video~\cite{singer2023makeavideo} and MagicVideo~\cite{zhou2022magicvideo} incorporate spatio-temporal attention (ST-Attn) to handle temporal aspects. Other recent methods focus on localizing edits within the video, such as Video-P2P~\cite{liu2023video}, which uses decoupled-guidance attention to ensure semantic consistency, and Pix2Video~\cite{ceylan2023pix2video}, which propagates anchor frame edits. 
\vspace{-1.5em}
\paragraph{Diffusion-Based 3D Editing.}
Recently, diffusion-based NeRF editing has garnered significant interest.  Instruct 3D-to-3D~\cite{kamata2023instruct} and Instruct-NeRF2NeRF (IN2N)~\cite{haque2023instruct} utilize Instruct-Pix2Pix (IP2P)~\cite{brooks2023instructpix2pix}, an image-conditioned diffusion model, to enable instruction-based 2D image editing.  Similarly, IN2N~\cite{haque2023instruct} proposes an Iterative Dataset Update (Iterative DU) technique that alternates between editing NeRF-rendered images using the diffusion model and updating the NeRF representation during training based on the edited images. ViCA-NeRF~\cite{dong2024vica} extends IN2N~\cite{haque2023instruct}, leveraging depth information from NeRF to propagate modifications in key views across other views, ensuring spatial consistency. DreamEditor~\cite{zhuang2023dreameditor} employs DreamBooth~\cite{ruiz2023dreambooth} as a 2D prior and uses SDS loss to facilitate precise text-driven editing. 
\vspace{-1.5em}

\paragraph{4D Scene Editing.}

Earlier 4D scene editing methods~\cite{qiao2022neuphysics,jiang20234d} remain limited in advanced, real-time editing capabilities. Recent advancements, such as Control4D~\cite{shao2023control4d} and Instruct-4D-to-4D (I4D-to-4D) \cite{mou2024instruct}, have made progress in improving consistency for 4D scene editing, yet these methods rely significantly on supplementary models beyond diffusion-based approaches. For instance, I4D-to-4D integrates a pre-trained optical flow model\cite{teed2020raft} to maintain temporal alignment across frames, while Control4D uses a GAN-based framework to manage dynamic adjustments and edits. However, this reliance on external models introduces notable limitations. GAN architectures are known for their instability, which can complicate training and reduce reliability during edits, while optical flow models may yield unreliable results in novel or complex scenes where accurate flow guidance is essential. Additionally, I4D-to-4D's dependency on an anchor-aware attention module means that the performance can vary based on the chosen anchor, introducing potential inconsistencies in editing results. Consequently, the added complexity of these external models often restricts the overall performance gains, highlighting the need for more robust and adaptable solutions in 4D editing. Our proposed framework solely focuses on controlling the diffusion process rather than relying on the performance of external models. 

% NeRFPlayer~\cite{nerfplayer} divides the 4D space into static, deforming, and newly updated regions based on temporal characteristics. Despite these advancements, a major limitation remains the lack of user-friendly editing capabilities for dynamic scenes. Current methods do not allow users to intuitively edit or modify these scenes, particularly when following specific instructions. This represents a key area for future research, where enhancing interactivity and user-driven editing capabilities could greatly increase the practicality of NeRF-based dynamic scene representations.

\begin{figure*}[t]
    \centering
    \includegraphics[width=0.95\linewidth]{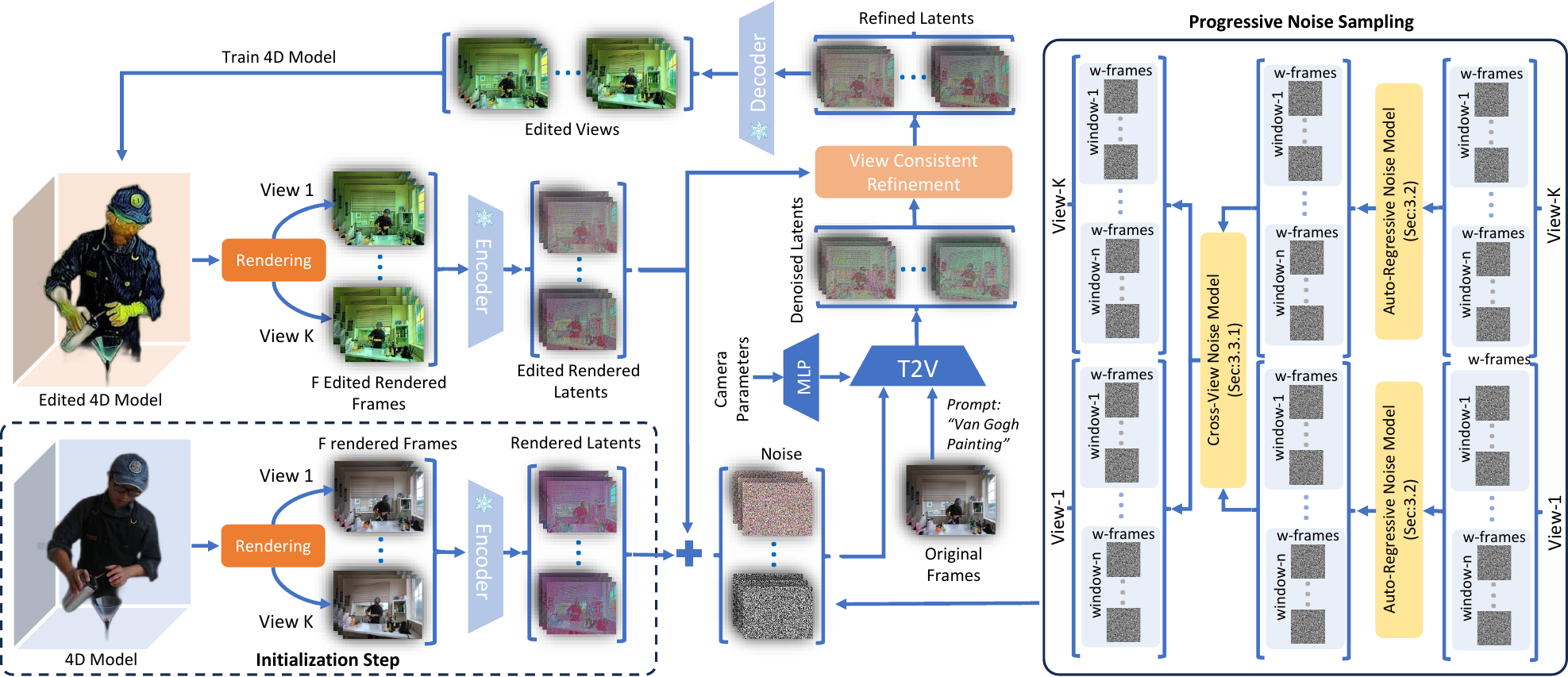}
    \vspace{-2mm}
   \caption{\footnotesize \textbf{PSF-4D framework} is designed for text-guided 4D editing. \textbf{Right.} We introduce a progressive noise sampling method for noise initialization, consisting of two key stages: \textbf{(i)} an \emph{autoregressive noise model} to ensure temporal consistency and \textbf{(ii)} \emph{cross-view noise control} to maintain spatial coherence. \textbf{Left.} This technique is incorporated into the diffusion process of the text-to-video (T2V) editing model, enabling the generation of 4D scenes with spatio-temporal coherence across multiple views. We further refine the edited 4D scene by enforcing a \emph{view-consistent refinement} strategy. Note that we consider this refinement process only after constructing the initial edited 4D model, i.e. $l >= 1$ (Sec.~\ref{sec:view_consistency}). After the initialization step ($l = 0$), we do not consider the original rendered latents anymore; only edited rendered latents have a role in next stages ($l >= 1$).}
    \label{fig:summary}
    \vspace{-3mm}
\end{figure*}

\vspace{-1mm}
\section{Methodology}
\vspace{-1mm}
In Figure~\ref{fig:summary}, we present our proposed framework, PSF-4D, where we achieve text-driven 4D editing using a 2D image diffusion model. Since a 4D scene consists of multi-view video data, we start with the adaptation of a T2I model to a T2V model that can perform multi-view video editing. However, there are two challenges associated with this adaptation: \emph{temporal consistency} and \emph{multi-view consistency}. To overcome these challenges, we propose a \emph{progressive noise sampling} strategy that consists of two noise initialization models: \emph{auto-regressive noise model (ANM)} to enforce temporal consistency and \emph{cross-view noise model (CNM)} to enforce multi-view consistency. We take additional measures for preserving multiview geometry information during editing: \textbf{i)} a \emph{view-consistent refinement} technique that iteratively refines the edits obtained from the T2V model; \textbf{ii)} \emph{view-aware positional encoding} to distinguish between different views.

\vspace{-1em}
\paragraph{Text-to-4D Editing.} Following Instruct4D-to-4D~\cite{mou2024instruct}, we consider a 4D scene as a pseudo-3D representation where each pseudo-view consists of a sequence of multiple frames in a video format. Given a text instruction \( C_T \) and a set of extrinsic camera parameters $\kappa \in \mathbb{R}^{K \times 16}$, our goal is to edit a multi-view video from K different view angles, \( I  \in \mathbb{R}^{K \times F \times H \times W \times C } \). Here, we can edit each view using a T2V model and then train a 4DGS model~\cite{wu20244d} on the edited views. For the T2V model, we take a pre-trained Stable Diffusion V2.1 (SD)~\cite{Rombach_2022_CVPR} image editing model and follow Tune-A-Video (TAV)~\cite{wu2023tune} to inflate the 2D convolutions to 3D convolutions. Similarly to SD, \textit{training} video diffusion models also consist of a diffusion process paired with a denoising process, both operating within the latent space of an autoencoder, $\mathcal{E}$. During forward diffusion, \emph{i.i.d} noise $\epsilon \sim \mathcal{N}(\mathbf{0},\mathbf{I})$  is added to the latent \( z = \mathcal{E}(I) \) to produce a noisy latent \( z_t \), with noise level set by a random timestep \( t \in T \). For reverse diffusion, we consider DDIM sampling~\cite{song2020denoising} process where we start from a latent \( z_T \) with maximum noise. The T2V model (with parameters $\theta$) is trained to predict the clean latent for the next timestep \( \tilde{z}_{t-1} \) as \( \tilde{\epsilon}_t = \tilde{\epsilon}_\theta (z_t, t, \kappa,  I, C_T)\). Here, the model is trained to approximate the noise added during the forward diffusion. The update rule for each timestep \( t \) is defined as:
\begin{equation}\label{eq:denoising}
    \tilde{z}_{t-1} = \sqrt{\alpha_{t-1}} \left( \frac{z_t - \sqrt{1 - \alpha_t} \, \tilde{\epsilon}_t}{\sqrt{\alpha_t}} \right) + \sqrt{1 - \alpha_{t-1}} \, \tilde{\epsilon}_t,
\end{equation}
where \( \alpha_t \) and \( \alpha_{t-1} \) control the noise level at each step. Please see \emph{Supplementary} for more details on 4DGS and the diffusion process. 

\subsection{Temporal Consistency}\label{sec:temp_consistency}
% Recent development of video diffusion models has made notable efforts to ob . To achieve better consistency for long videos, we propose the following noise initialization scheme. 
% % For example, methods like TAV~\cite{wu2023tune}, Video-P2P~\cite{liu2023video}, FateZero~\cite{qi2023fatezero} work well for short videos (e.g. 8 frames) where we are dealing with longer videos. 

% % To introduce such correlations in the noise across frames in a video diffusion process, we start by modifying the standard assumption of uncorrelated Gaussian white noise.
% \vspace{-1em}
% \paragraph{Auto-regressive Noise Model}  
In case of \emph{i.i.d}, noise across frames is drawn independently from a Gaussian distribution \( \epsilon \sim \mathcal{N}(\mathbf{0}, \mathbf{I}) \), where \( \mathbf{I} \) represents the identity covariance matrix. Since this independent frame-wise noise model does not consider cross-frame correlations, the generated video may contain inconsistent or jittery frame transitions. Therefore, we replace this independent noise assumption with a correlated noise sequence generated by an autoregressive (AR) model. 
% Specifically, an AR process can be applied across frames, where the noise in each frame is conditioned on the noise of the previous frame.% where each \( {\epsilon}^t \in \mathbb{R}^d \) denotes the noise component at frame \( t \), and \( d \) represents spatial dimensions.
To this end, we take a window-based approach where we have \( n \) number of windows with each having \( w \) frames, i.e. $F= nw$. Let the noise tensor \( {\epsilon} = ({\epsilon}^1, {\epsilon}^2, \dots, {\epsilon}^n)^{\top} \) represent the noise values across \( n \) windows. We define the AR(1) model as follows:
\begin{equation}\label{eq:pnm}
    {\epsilon}^i = \gamma {\epsilon}^{i-1} + \sqrt{1 - \gamma^2} \, \boldsymbol{\eta}_i, \quad \boldsymbol{\eta}_i \sim \mathcal{N}(\mathbf{0}, \mathbf{I}),
\end{equation}
where \( \gamma \in (0, 1) \) is a parameter controlling the degree of temporal correlation between consecutive windows, and \( \boldsymbol{\eta}_i \) represents independent Gaussian noise at each window. 
\vspace{-1em}
\paragraph{Takeaways.} In above model, noise is destructively added in the correlated manner (mimicking realistic motion), the noisy video data (clean video + correlated noise) more closely resembles the real distribution of possible corrupted videos. The advantage is that it has become easier now for the network to perform reverse mapping because it better matches real-world video dynamics. Therefore, by explicitly modeling correlation, we reduce the mismatch between forward noising and real video motion, mitigating the chance of producing flickering frames.

\subsection{Multi-View Consistency}\label{sec:view_consistency}
\vspace{-1mm}
After performing temporally consistent editing in all views, we can train a 4D model on the given edited views. However, simply regenerating these edits again and again still produces inconsistent results due to the issue of multiview consistency. Hence, we propose to enforce multi-view consistency in the final editing through: \emph{Cross-view Noise Model} and \emph{View Consistent Refinement (VCR)}. 
\vspace{-0.5mm}
\subsubsection{Cross-View Noise Model}\label{sec:CNC} 
% \vspace{-1mm}
Although the auto-regressive noise model is better suited for temporal consistency, spatial coherence across different perspectives is more important in multi-view generation. Therefore, we consider a slightly different noise model in this case. Considering $K$ views of a 4D scene, where each view is a video of $n$ windows. Before considering the crows-view noise model, we first apply the auto-regressive noise model to all $K$ views $\boldsymbol{\epsilon} = \{ {\epsilon}^i_k \}$ where $i \in [1,n]$ and $k \in [1,K]$. Here, ${\epsilon}^i_k$ represents the noise value for \( i^{\text{th}} \) window of the \( k^{\text{th}} \) view. 

For better understanding, we present a window-by-window noise model as the T2V framework processes one window of a specific view at a time.  For \( i^{\text{th}} \) window of all $K$ views, let \( \hat{{\epsilon}}^i = (\hat{\epsilon}^i_{1}, \hat{\epsilon}^i_{2}, \dots, \hat{\epsilon}^i_{K})^{\top} \) denote the tensor comprising noise components for individual views. Here, \( \hat{\epsilon}^i_{k} \) corresponds to the \( k^{\text{th}} \) element in the noise tensor \( \hat{\epsilon}^i \). We introduce a shared noise component \( \hat{{\epsilon}}^{i}_{\text{shared}} \) that is constant across all views, establishing a baseline level of similarity among the generated views. This is crucial in multi-view generation, where maintaining coherence across different perspectives of the same scene is essential for realistic rendering. On the other hand, another component \( \hat{\epsilon}^i_{k, \text{ind}} \) is also considered, which provides the individual noise for each view. Adding \( \hat{\epsilon}^i_{k, \text{ind}} \) allows for controlled variation across views, capturing slight differences that would naturally occur when observing a 3D object or scene from multiple angles. This helps avoid rigid or overly uniform results that can occur when only shared noise is applied. The final noise for each view \( \hat{\epsilon}_i^{k} \) is then constructed as a linear combination of these two components.
\begin{align}\label{eq:cnm} 
    \hat{{\epsilon}}^{i}_{\text{shared}} &\sim \mathcal{N}\left(\mathbf{0}, \lambda\mathbf{I} \right), \quad \hat{\epsilon}^i_{k, \text{ind}} = \sqrt{1 - \lambda}{\epsilon}^i_k, \\
    \hat{\epsilon}^i_{k} &= \hat{{\epsilon}}^{i}_{\text{shared}} + \hat{\epsilon}^i_{k, \text{ind}} \nonumber
\end{align}
Here, $\lambda \in (0, 1)$  controls the balance between shared and individual noise contributions. This design introduces cross-view correlation through \( \hat{{\epsilon}}^{i}_{\text{shared}} \), while \( \hat{\epsilon}^i_{k, \text{ind}} \) retains unique noise characteristics for each view, creating both shared and individual noise features to enhance coherence and diversity in multi-view generation. Finally, we have the noise values for all windows and all views, $\hat{\boldsymbol{\epsilon}} = \{ {\hat{\epsilon}}^i_k \}$ where $i \in [1,n]$ and $k \in [1,K]$.

\begin{figure*}[t]
    \centering
    \includegraphics[width=0.925\linewidth]{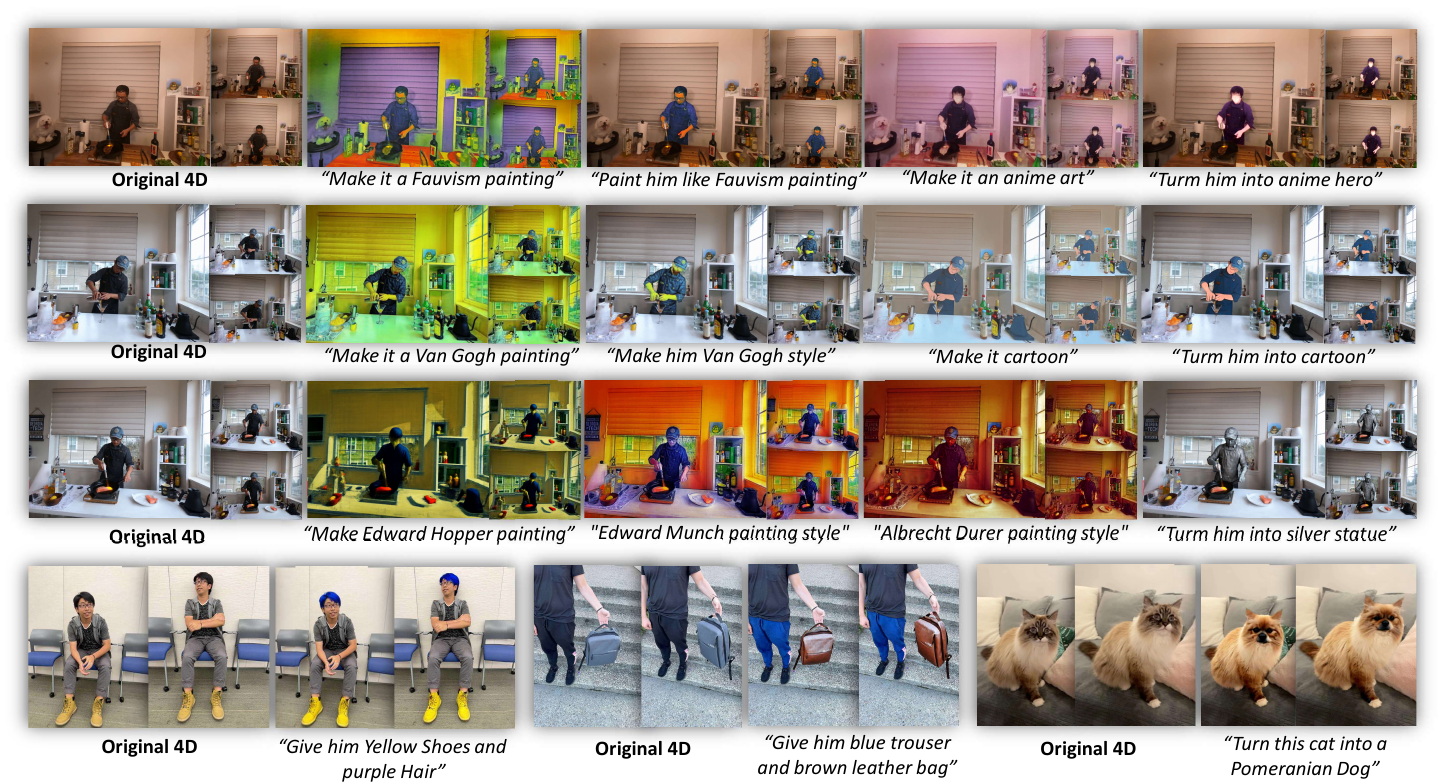}
    \vspace{-3mm}
    \caption{\footnotesize \textbf{Qualitative 4D Editing Results.} Examples of 4D editing tasks performed using our PSF-4D framework. Each row represents a specific editing scenario, demonstrating the versatility and precision of PSF-4D across a variety of tasks, including style transfer, object transformation, and attribute modification. From transforming a scene into different artistic styles (e.g., "Make it a Fauvism painting," "Make him Van Gogh style") to specific object edits (e.g., "Give him blue trousers and brown leather bag"), PSF-4D maintains consistency and coherence across frames in dynamic 4D scenes.}
    \vspace{-2mm}
\label{fig:qualitative_4d_editing}
\end{figure*}
Figure~\ref{fig:summary} illustrates the process of generating the initial set of edited views. Starting from the unedited 4D model, we render multiple views, which are passed through the VAE encoder to obtain their corresponding unedited latent representations. Using \( \hat{\boldsymbol{\epsilon}} \), we introduce noise to these latents, preparing them for processing through the Text-to-Video (T2V) model, conditioned on both text prompts and the original view information. After denoising, the latents are decoded through the VAE decoder, resulting in the initial edited views \( \tilde{I}_{0} = \{\tilde{I}_{0}^{k}\}, k \in [1, K] \). By training on this initial set \( \tilde{I}_{0} \), we obtain the initial edited 4D model. 

\vspace{-1em}
\paragraph{Takeaways.} In general, exact pixel-level or feature-level correspondences are traditionally used in classical geometry pipelines. However, a diffusion-based framework can learn the alignment implicitly if provided a suitable correlation prior. Our CNM is precisely this prior, bridging the gap between i.i.d. noise (which fails at multi-view consistency) and heavy explicit alignment. By incorporating 3D multi-view constraints at the noise level, view consistency naturally emerges in the reverse diffusion process as the diffusion process must jointly reconstruct shared structures and accommodate local variations. While progressive noise sampling helps achieve smooth motion, spatial coherence across views remains challenging with CNM-based noise control alone, potentially leading to minor inconsistencies or artifacts. To address this, we apply a view-consistent refinement technique with a focus on enhancing the spatial and temporal coherence of the edited 4D model.

\subsubsection{View Consistent Refinement}~\label{sec:view_refinement} 
Let us denote $\tilde{I}_{l}^{k}$ as \( k^{\text{th}} \) renderings of the edited 4D model at the \( l^{\text{th}} \) iteration of the refining process. We then obtain $\tilde{z}_{l}^{k}$ as the latent equivalent of $\tilde{I}_{l}^{k}$. After adding noise to the $\tilde{z}_{l}^{k}$, we use the T2V model with conditioning for denoising. Following Eq.~\ref{eq:denoising}, the denoised latent $\tilde{z}^{k}_{l+1}$  can be estimated after $T$ number of DDIM sampling steps. If we decode $\tilde{z}^{k}_{l+1}$ to $\tilde{I}_{l+1}^{k}$, we should have a higher quality view generation with smooth motion as compared to $\tilde{I}_{l}^{k}$. However, the same cannot be said for spatial consistency among views as the diffusion process of T2V struggles with view-consistent generation (even with the utilization of CNM). Therefore, we explicitly inject view information into the T2V editing pipeline. To this end, the rectified $\hat{z}^{k}_{l+1}$ are computed by 
\begin{equation}
    \hat{z}^{k}_{l+1} = \omega_l \tilde{z}_{l+1}^{k} + (1-\omega_l) \tilde{z}^{k}_{l},
    \label{eq:view_refinement}
\end{equation}
% trim=3.3cm 6.5cm 3.5cm 6.5cm, clip

Here, \( \omega_l \) is a predefined weight to balance between the denoising results \( \tilde{z}^{k}_{l+1} \) and the rendered multi-view consistent \( \tilde{z}_{l}^{k} \). The parameter \( \omega_l \) determines how much multi-view consistency is imposed on the denoising process.  Training a 4D model with more focus on \( \tilde{z}_{l}^{k} \) (low $\omega_l$) forces multi-view consistency but may oversmooth some regions. On the other hand, directly utilizing the denoising directions from \( \tilde{z}^{k}_{l+1} \) ($\omega_l = 1$) produces videos with more details but less multi-view consistency. At the beginning of the refinement stage, we focus more on the fidelity of the generated views while emphasizing more on the multi-view consistency at later iterations. Therefore, we start with a high value of \( \omega_l \) and slowly decrease it as the refinement progresses.  We repeat the refinement process for \( L \) steps. 
\subsubsection{View-Aware Position Encoding}
In our work, we fine-tune a T2V model with multi-view video data. To distinguish between different views while fine-tuning, view-aware position encoding is necessary which can be derived from the absolute camera parameters. To this end, we encode the camera parameters $\kappa$ by employing a 2-layer MLP with parameters $\Phi$ and add the resulting camera embeddings to time embeddings as residuals~\cite{shi2023mvdream}. Doing so provides additional view awareness to the T2V model and reduces spatial artifacts. 
\begin{figure*}[t]
    \centering
    \includegraphics[width=0.925\linewidth]{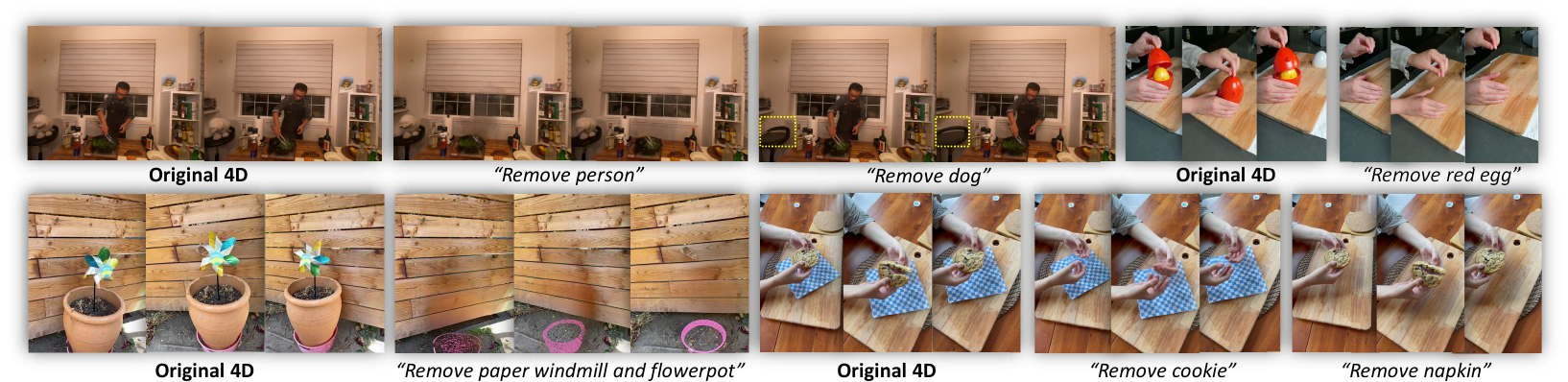}
    \vspace{-2mm}
   \caption{\footnotesize \textbf{Object Removal in 4D Scenes.} Examples of object removal across various scenes from different datasets, including DyNeRF, DyCheck, and HyperNeRF. Each row illustrates the original 4D scene followed by frames with specific objects removed, as per the editing prompt. Prompts such as “Delete Person", “Delete dog", “Remove paper windmill and flowerpot,” “Remove red egg", “Remove cookie", and “Remove napkin” demonstrate the capability of our method to accurately and seamlessly edit out targeted objects while preserving the surrounding scene consistency.}
    \vspace{-2mm}
    \label{fig:object_removal}
\end{figure*}
\begin{figure*}[t]
    \centering
    \includegraphics[width=0.915\linewidth]{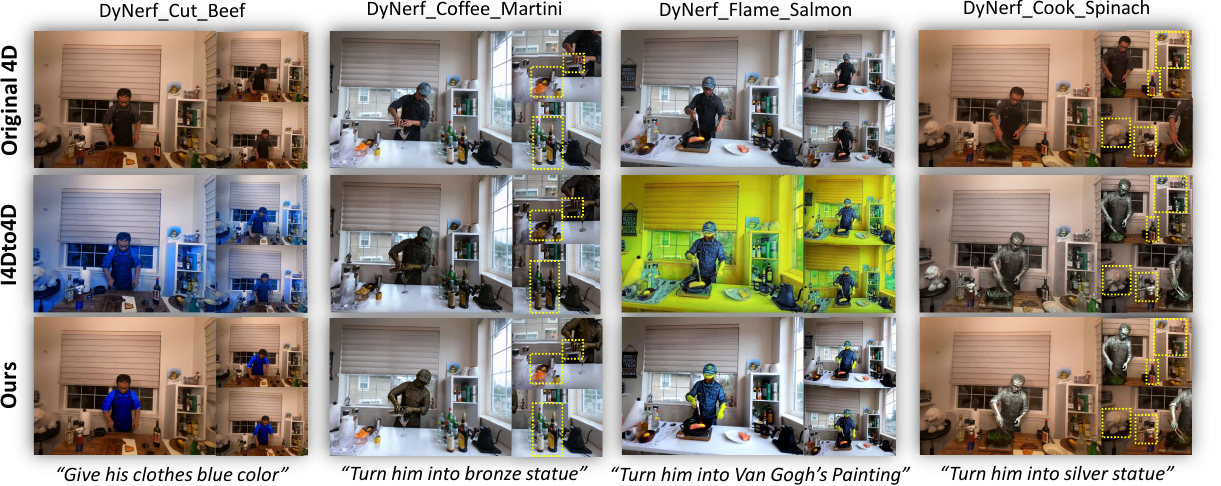}
    \vspace{-2mm}
    \caption{\footnotesize \textbf{Comparison of 4D Editing Results.} Examples of 4D scene editing using our approach compared to Instruct 4D-to-4D (I4Dto4D) and the original 4D scene across various scenes in the DyNeRF dataset. Each column presents a different editing prompt: \textit{DyNeRF\_Cut\_Beef} (“Give his clothes blue color”), \textit{DyNeRF\_Coffee\_Martini} (“Turn him into bronze statue”), \textit{DyNeRF\_Flame\_Salmon} (“Turn him into Van Gogh’s Painting”), and \textit{DyNeRF\_Cook\_Spinach} (“Turn him into silver statue”). The original 4D scenes (top row) show unedited content, while the I4D-to-4D~\cite{mou2024instruct} results (middle row) illustrate partial modifications. Our approach (bottom row) achieves more precise and consistent adherence to the editing prompts across all frames, producing visually coherent and realistic transformations.}
    \vspace{-2mm}
    \label{fig:comparison}
\end{figure*}

% At each iteration, the refined renderings \( \hat{z}^{k}_{l} \) are employed to refine the edited 4D field that improves the rendering quality. It needs to be emphasized that above refinement process is crucial in obtaining our desired 4D editing.   

\subsection{Training Objective}
For scene-specific adaptation of the T2V model, we minimize the following multi-view diffusion loss,
\begin{align}
\label{eq:cond_ldm_loss}
\mathcal{L}(\theta, \Phi) = \mathrm{E}_{I, C_T , \kappa, t, \hat{\boldsymbol{\epsilon}}} \Big[
\Vert \hat{\epsilon} - \tilde{\epsilon}_\theta (z_t, t, \kappa,  I, C_T) \Vert_{2}^{2}.
\Big]
\end{align}
In our work, we tune the model for around 3000 iterations on each scene before utilizing it for text-guided editing.

\section{Experiments}
\label{sec:experiments}
Our implementation is built on the PyTorch framework and tested on a single RTX A6000 GPU, utilizing the 4D Gaussian Splatting framework~\cite{wu20244d} for constructing 4D scenes. The model initialization phase involves 12,000 training iterations, 10000 iterations in the coarse phase to optimize static 3D Gaussian, and 2000 iterations in the fine phase to refine 4D Gaussian. For example, in the HyperNeRF  dataset~\cite{park2021benchmark}, we use a rendering resolution of 960×540, achieving a rendering speed of 34 FPS. The editing phase adds 10000 more iterations. Considering the fine-tuning of the T2V model, PSF-4D has a total training time of approximately 4 hours. In addition, we incorporate SAM~\cite{ravi2024sam} to achieve local and precise editing.
\vspace{-1em}
\paragraph{Datasets.}  
We evaluate our method on 4D scenes captured using both single hand-held cameras and multi-camera setups. These include: (I) \textit{Monocular} scenes from the DyCheck \cite{gao2022monocular} and HyperNeRF \cite{park2021hypernerf} datasets, featuring object-centric scenes with single moving cameras, and (II) \textit{Multi-camera} scenes from DyNeRF/N3DV \cite{li2022neural}, consisting of indoor environments with human motions and multiple camera perspectives. 
\vspace{-1em}
\paragraph{Baselines.} 
We compare our approach with the baseline I4D-to-4D~\cite{mou2024instruct} and an extended version of the 3D editing framework, IN2N~\cite{haque2023instruct}, which we adapted into a 4D variant (IN2N+4D) by iteratively editing each frame and incorporating it back into the dataset. For quantitative evaluation, we utilize the Fréchet Video Distance (FVD)~\cite{unterthiner2018towards} and Fréchet Inception Distance (FID)~\cite{heusel2017gans} metrics to assess the visual similarity between the edited dataset and generated images. Additionally, we calculate the CLIP cosine similarity (CLIP-S)~\cite{radford2021learning} to measure alignment between generated images and textual descriptions, thereby providing a robust evaluation of both visual fidelity and semantic relevance in the edits. We also consider other performance metrics such as peak signal-to-noise ratio (PSNR), SSIM, and LPIPS. Details are in \emph{Supplementary}.

% \vspace{-1mm}

% \vspace{-1mm}
\subsection{Qualitative Evaluation}
\label{sec:qualitative_evaluation}
% \vspace{-1mm}
Our PSF-4D framework demonstrates robust 4D editing capabilities across four key tasks: 
% multi-attribute editing, object removal, style transfer, and local editing. Each of these editing scenarios showcases the versatility and precision of PSF-4D in handling complex modifications within 4D scenes.
\vspace{-1em}
\paragraph{Multi-Attribute Editing.}
In multi-attribute editing, PSF-4D effectively manages multiple modifications on a single subject. For example, in Figure~\ref{fig:qualitative_4d_editing}, the prompt “Give him blue trousers and a brown leather bag” requires simultaneous color and object modifications. PSF-4D successfully applies both edits consistently across frames, demonstrating its ability to handle compound changes without compromising coherence.

\vspace{-1em}
\paragraph{Style Transfer.} PSF-4D excels in style transfer tasks, as seen in Figure~\ref{fig:qualitative_4d_editing} with prompts like “Make it a Van Gogh painting” and “Turn him into silver statue”. PSF-4D accurately applies the specified artistic styles across the entire scene, maintaining consistency in both spatial and temporal dimensions. The stylistic transformations are visually coherent, reflecting PSF-4D’s superior control in scene-wide aesthetic changes.

\vspace{-1em}
\paragraph{Object Removal.} Figure~\ref{fig:object_removal} highlights PSF-4D's capability in object removal tasks. Prompts such as “Remove person,” “Remove red egg,” and “Remove napkin” illustrate how PSF-4D seamlessly removes targeted objects from complex scenes while preserving background integrity and spatial consistency. This precise control in object manipulation highlights PSF-4D’s advanced scene understanding in dynamic 4D environments.

\begin{table}[t]
\centering
\caption{\footnotesize \textbf{Quantitative Comparison} across 100 dynamic scene edits.}
\vspace{-2mm}
\label{tab:quantitative_comparison}
\scalebox{0.65}{\begin{tabular}{l|c|c|c|c|c}
\toprule
\textbf{Method} & \textbf{FVD} $\downarrow$ & \textbf{FID} $\downarrow$ & \textbf{CLIP-S} $\uparrow$ & \textbf{PSNR} $\uparrow$ & \textbf{SSIM} $\uparrow$ \\
\midrule

IN2N~\cite{haque2023instruct}+HexPlane~\cite{hexplane} & 382.6 & 68.75 & 0.2985 & 17.28 & 0.652 \\
% Tensor4D & 155.6 & 0.3144 \\
% Tensor4D+GAN & 47.39 & 0.3178 \\
I4D-to-4D~\cite{mou2024instruct} & 294.1 & 37.58 & 0.3045 & 19.74 & 0.697 \\
PSF-4D (Ours) & \textbf{215.3} & \textbf{25.39} & \textbf{0.3292} & \textbf{21.85}  & \textbf{0.728}\\
\bottomrule
\end{tabular}}
\vspace{-3mm}
\end{table}

% In addition to using SAM, we do the following. Given a 4D scene, let us randomly choose a view \( I \). We use a widely used vision language model, LlaVA~\cite{li2023m}, to generate a detailed description \( C_D \). We feed \( C_D \) and a single-sentence edit instruction \( C_T \) to an LLM (e.g. GPT-4) to get the summarized edit instruction \( C_S \) that encodes crucial editing and content information of \( I \). Finally, we use \( C_S \) as the final editing prompt for T2V. This results in fine-grained precision, i.e. applying changes to specific parts of an object while keeping the rest of the scene unaffected. For instance, in Figure~\ref{fig:comparison}, the prompt \( C_T = \) “Give his clothes blue color” requires targeted color alteration to only the clothing, which PSF-4D accomplishes with high accuracy and without unintended modifications to other areas. This level of selectivity underscores PSF-4D’s capability for detailed, localized edits within complex 4D scenes. We explain more on this in \emph{Supplementary}.

\vspace{-1em}
\paragraph{Local Editing.} Figure~\ref{fig:comparison} demonstrates the local editing capabilities of our proposed approach. In addition to using SAM, we apply a special type of prompt engineering to obtain superior results. We explain more in \emph{Supplementary}.

Qualitative results in different editing tasks emphasize PSF-4D’s adaptability and precision in 4D scene editing, consistently outperforming baseline methods by producing high-quality, contextually aligned modifications.

\subsection{Quantitative Evaluation}
\vspace{-1mm}
\label{sec:quantitative_evaluation}

\begin{figure}[t]
    \centering
    \includegraphics[width=0.6\linewidth, trim={7cm 7cm 7cm 7cm}, clip]{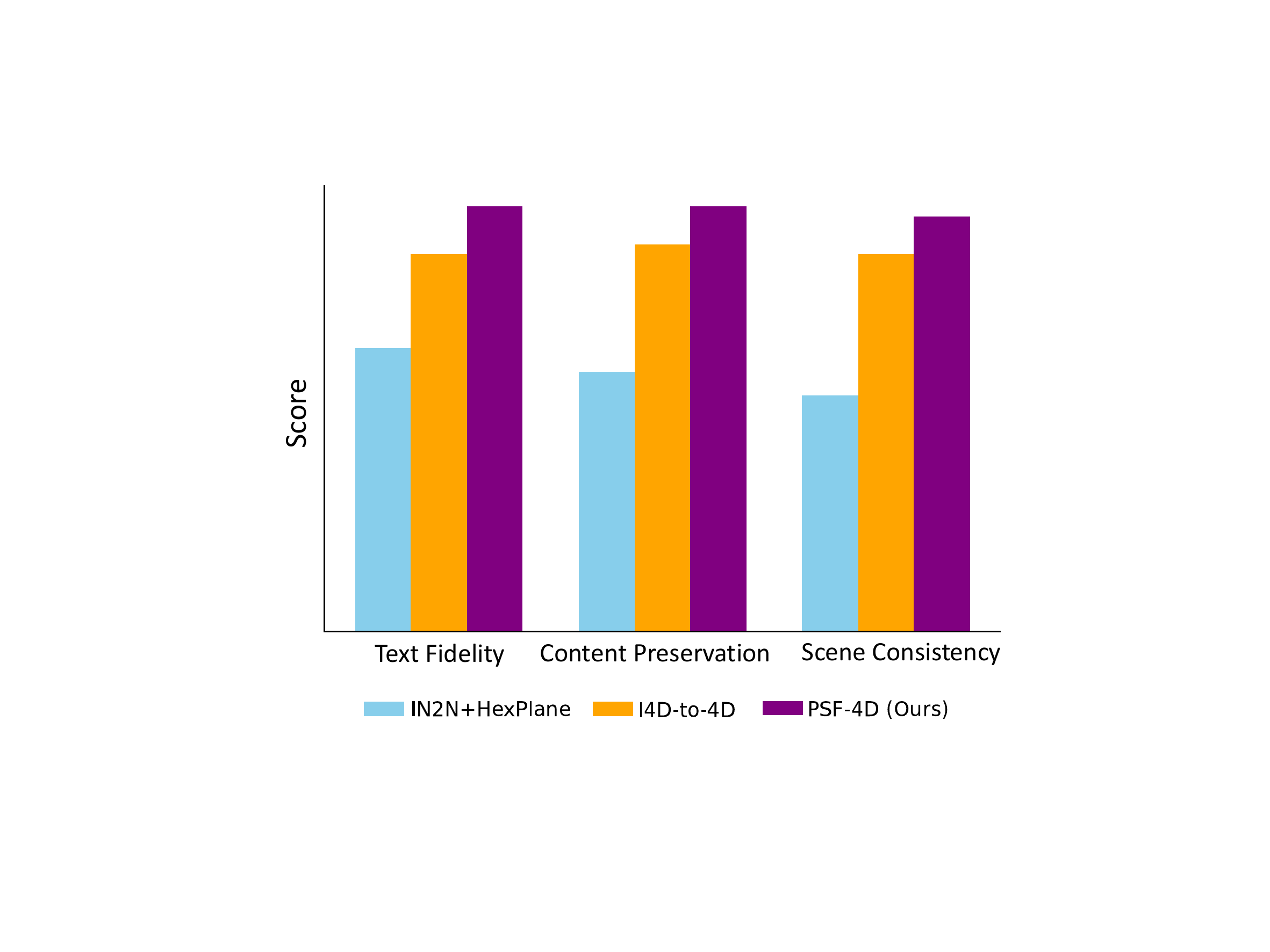}
    \vspace{-1mm}
    \caption{\footnotesize \textbf{User Study Evaluation}. Results of a user study comparing text fidelity, content preservation, and scene consistency across three methods: IN2N+HexPlane, I4D-to-4D, and our proposed PSF-4D. PSF-4D demonstrates superior performance across all criteria, with particularly high scores in text fidelity and scene consistency, indicating its effectiveness in producing accurate, coherent edits in dynamic scenes.
 }
    \label{fig:user_study}
    \vspace{-1mm}
\end{figure}

The quantitative results in Table~\ref{tab:quantitative_comparison} demonstrate the effectiveness of our proposed PSF-4D framework compared to existing methods on 100dynamic scene edits. We evaluate the methods using Frechet Inception Distance (FID) to measure the quality of generated images and CLIP Similarity to assess alignment with the textual prompts. Our PSF-4D approach achieves the lowest FID score of 20.39, indicating superior visual quality and coherence in the edited outputs compared to IN2N+HexPlane~\cite{hexplane} and I4D-to-4D~\cite{mou2024instruct}, which have FID scores of 68.75 and 37.58, respectively. Additionally, PSF-4D attains the highest CLIP Similarity score of 0.3292, reflecting better alignment with the intended editing prompts than the other methods. These results highlight PSF-4D's ability to produce contextually accurate and visually realistic edits, outperforming baseline methods in both perceptual quality and prompt adherence.
\vspace{-1mm}
\paragraph{User Study.} For the user study shown in Figure~\ref{fig:user_study}, we surveyed a random sample of 100 participants aged between 21 and 40. Participants were asked to rate the generated edits based on three key aspects: text fidelity, content preservation, and scene consistency. Scores were then averaged for each method, with PSF-4D consistently achieving the highest ratings across all categories, reflecting strong user preference and perceived quality of edits.

\begin{table}[t]
\centering
\caption{\footnotesize \textbf{Ablation Study} on different components of PSF-4D: \emph{auto-regressive noise model (ANM), cross-view noise model (CNM), view consistent refinement (VCR)}, and \emph{view-aware position encoding (VPE)}.  We consider the DyNeRF dataset with 8 different prompts for this experiment.}
\vspace{-2mm}
\label{tab:quantitative_comparison_ablation}
\scalebox{0.635}{\begin{tabular}{l|c|c|c|c|c|c}
\toprule
\textbf{Method} & \textbf{FVD} $\downarrow$ & \textbf{FID} $\downarrow$ & \textbf{CLIP-S} $\uparrow$ & \textbf{PSNR}$\uparrow$ & \textbf{SSIM}$\uparrow$ & \textbf{LPIPS}$\downarrow$ \\
\midrule
IN2N~\cite{haque2023instruct}+HexPlane~\cite{hexplane} & 379.2 & 64.27 & 0.2971 & 16.71 & 0.649 & 0.374 \\
% Control4D~\cite{shao2023control4d} & 49.13 & 0.2980 & 17.60 & 0.658 & 0.388 \\
I4D-to-4D~\cite{mou2024instruct} & 281.5 & 34.52 & 0.3068 & 19.92 & 0.706 & 0.419 \\
\midrule
PSF-4D \emph{w/o VCR} & 290.4 & 39.84 & 0.2941 & 17.36 & 0.673 & 0.397  \\
PSF-4D \emph{w/o CNM} & 262.8 & 33.06 & 0.2994 & 19.84 & 0.692 & 0.418 \\
PSF-4D \emph{w/o ANM} & 243.7 & 28.17 & 0.3078 & 20.96 & 0.714 & 0.427 \\
PSF-4D \emph{w/o VPE} & 229.1 & 26.12 & 0.3104 & 21.15 & 0.718 & 0.430 \\
PSF-4D & \textbf{210.4} & \textbf{22.58} & \textbf{0.3241} & \textbf{22.17} & \textbf{0.726} & \textbf{0.436}  \\
\bottomrule
\end{tabular}}
\vspace{-3mm}
\end{table}

% \vspace{-2mm}
\subsection{Ablation Study}
\begin{figure}[t]
    \centering
    \includegraphics[width=0.9\linewidth]{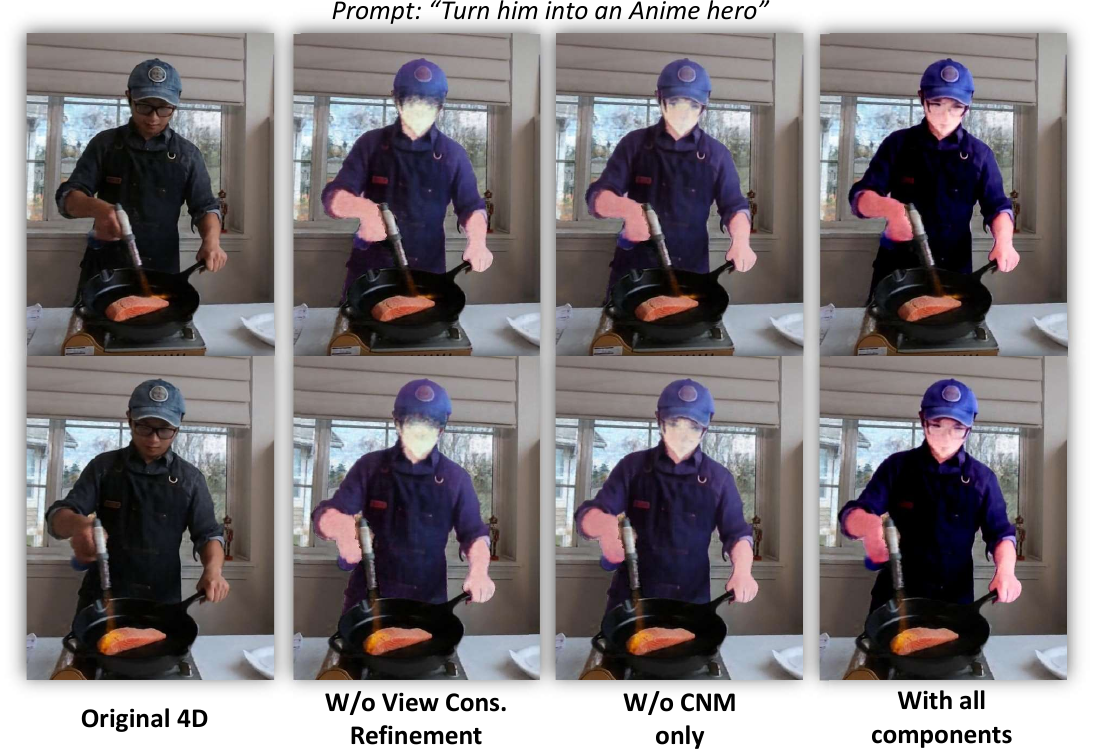}
    \vspace{-3mm}
    \caption{\footnotesize \textbf{Ablation on different components of our proposed method. } We show the impact of CNM and view consistent refinement in obtaining the desired editing effect. PSF-4D w/o view consistent refinement (VCR) indicates we edit the model only once ($L=0$), resulting in poor editing. Without CNM, the initial edited 4D model's quality drops significantly. However, quality improvement can be observed with iterative VCR. These results suggested that VCR plays the most important role in achieving the SOTA 4D editing performance.}
    \vspace{-3mm}
    \label{fig:main_ablation}
\end{figure}

\vspace{-1mm}
We study the impact of different hyperparameters on the performance of PSF-4D. Figure~\ref{fig:main_ablation} and Table~\ref{tab:quantitative_comparison_ablation} show the impact of different components on the overall performance of PSF-4D. It can be observed that all of these components play important roles in obtaining our desired results. For choosing the values for different hyperparameters, we also conduct further ablation. For instance, we choose $\lambda = 0.7$ and $\gamma = 0.65$ to obtain the best editing performance. On the other hand, we start with $\omega_1 = 0.9$ and decrease it to $\omega_L = 0.6$ at the end of the refinement stage. Due to the page constraints, details of these choices along with other types of studies have been included \emph{Supplementary}.

\section{Discussion and Limitations}
Although we followed the Tune-a-Video framework and used stable diffusion image editing model, PSF-4D is designed to be independent of the underlying image editing model and T2V framework, allowing it to be readily integrated with various off-the-shelf models to achieve desired editing outcomes. This modular approach ensures the broad applicability and flexibility of our framework.   

While PSF-4D leverages progressive noise modeling to maintain temporal and view consistency, its reliance on noise control may lead to suboptimal results in highly dynamic or complex scenes where noise-based conditioning alone is insufficient to fully capture intricate spatial-temporal relationships. Although we introduce view-consistent refinement to overcome the shortcomings of our proposed noise modeling, it may inadvertently lead to over-smoothing of fine details. This can reduce the realism of textured or highly detailed objects in the 4D scene, particularly when too many refinement steps are applied.  Choosing $\omega_l$ and $L$ properly may prevent this. In addition, an adaptive noise control that dynamically adjusts $\gamma$ and $\lambda$ based on scene complexity could improve PSF-4D’s handling of diverse or highly dynamic content.

\vspace{-1mm}
\section{Conclusion}\label{sec:conclusion}
\vspace{-0.5mm}
PSF-4D offers an effective framework for achieving consistent and high-quality 4D video editing. By combining progressive noise sampling with iterative refinement, PSF-4D addresses key challenges in maintaining both temporal and view coherence across frames and perspectives. Leveraging autoregressive noise initialization and a cross-view noise model, the framework captures temporal dependencies and spatial alignment, while view-consistent iterative refinement ensures precise and stable edits. PSF-4D demonstrates a robust approach for complex 4D editing tasks, laying the groundwork for future improvements in efficiency and adaptability for dynamic scene editing across various applications. Diverse editing capabilities in multiple benchmarks have demonstrated the merit of our proposed framework.
% \vspace{-1em}
{
    \small
    \bibliographystyle{ieeenat_fullname}
    \bibliography{main}
}

\footnotetext[1]{Equal Contribution}
\end{document}